\documentclass{ifacconf}

\usepackage{graphicx}      
\usepackage{natbib}        
\usepackage{color}

\usepackage{amsmath} 
\usepackage{amssymb}  

\usepackage{algorithm} 
\usepackage[noend]{algpseudocode} 

\DeclareMathOperator*{\argmin}{argmin}

\definecolor{gray}{RGB}{200,200,200}
\definecolor{orange}{RGB}{237,125,49}
\newcommand{\vecb}[1]{\textbf{\textit{#1}}} 
\algnewcommand{\Initialize}[1]{%
  \State \textbf{initialize} {\raggedright #1}
}

\makeatletter
\let\old@ssect\@ssect 
\makeatother
\usepackage{natbib}
\usepackage{hyperref}
\hypersetup{
    colorlinks=true,
    linkcolor=blue,
    filecolor=magenta,      
    urlcolor=cyan,
    pdftitle={Overleaf Example},
    pdfpagemode=FullScreen,
    }
\makeatletter
\def\@ssect#1#2#3#4#5#6{%
	\NR@gettitle{#6}
	\old@ssect{#1}{#2}{#3}{#4}{#5}{#6}
}
\makeatother
\begin{document}
\begin{frontmatter}

\title{Learning Flight Control Systems from Human Demonstrations and Real-Time Uncertainty-Informed Interventions\thanksref{footnoteinfo}} 

\thanks[footnoteinfo]{Research was sponsored by the U.S. Army Research Laboratory and was accomplished under Cooperative Agreement Number W911NF-18-2-0134. The views and conclusions contained in this document are those of the authors and should not be interpreted as representing the official policies, either expressed or implied, of the Army Research Laboratory or the U.S. Government.}

\author[First]{Prashant Ganesh} 
\author[First]{J. Humberto Ramos} 
\author[Second]{Vinicius G. Goecks}
\author[First]{Jared Paquet}
\author[First]{Matthew Longmire}
\author[Second]{Nicholas R. Waytowich}
\author[Third]{Kevin Brink}

\address[First]{Department of Mechanical and Aerospace Engineering, University of Florida REEF, Shalimar, FL 32579, USA (e-mails: prashant.ganesh@ufl.edu, 	jramoszuniga@ufl.edu, jaredpaquet@ufl.edu, m.longmire@ufl.edu)}
\address[Second]{Human Research \& Engineering Directorate, U.S. Army Research Laboratory, Aberdeen Proving Ground, MD 21005, USA (e-mails: vinicius.goecks@gmail.com, nicholas.r.waytowich.civ@army.mil)}
\address[Third]{Munitions Directorate, Air Force Research Lab, Eglin AfB, FL 32542, USA (e-mail: kevin.brink@us.af.mil)}

\begin{abstract}                
This paper describes a methodology for learning flight control systems from human demonstrations and interventions while considering the estimated uncertainty in the learned models.
The proposed approach uses human demonstrations to train an initial model via imitation learning and then iteratively, improve its performance by using real-time human interventions. The aim of the interventions is to  correct undesired behaviors and adapt the model to changes in the task dynamics.
The learned model uncertainty is estimated in real-time via Monte Carlo Dropout and the human supervisor is cued for intervention via an audiovisual signal when this uncertainty exceeds a predefined threshold.
This proposed approach is validated in an autonomous quadrotor landing task on both fixed and moving platforms. It is shown that with this algorithm, a human can rapidly teach a flight task to an unmanned aerial vehicle via demonstrating expert trajectories and then adapt the learned model by intervening when the learned controller performs any undesired maneuver, the task changes, and/or the model uncertainty exceeds a threshold.
\end{abstract}


\end{frontmatter}

\section{Introduction}
The design and implementation of analytical control laws can be intractable or prohibitively challenging for many of the capabilities desired from autonomous or intelligent systems. This is precisely the case for highly complex systems that are extremely difficult to model. 
In these situations, machine learning-based methods can provide a useful alternative to formulating a classic controller to achieve intricate capabilities and capture complex dynamic behaviors.
Although machine learning techniques often lack analytical guarantees, they can produce highly effective controllers, also called policies, to perform challenging tasks with relatively minimal first-principles design and tuning.
Previous works have demonstrated that such techniques can be applied to embodied hardware platforms \cite{Ng2006,Ross2013,Giusti2015,Zhang2016,Bojarski2016,Gandhi2017,Goecks2018}, simulated \cite{Gu2017ICRA,Nair2018ICRA} and physical robots \cite{OpenAI2018,Tobin2018IROS,Thomas2018ICRA}, including aerial vehicles \cite{Blukis18, Anderson2018, Mei2016AAAI, Misra2017EMNLP, Tai2017IROS}.


These end-to-end data-driven approaches often use deep neural networks to represent the control policies, which require large amounts of data, and thus, lengthy training time to achieve the desired system behaviors. 
In addition, poorly-performing policies prior to convergence to desired behaviors can lead to catastrophic actions that may not be tolerable when working with physical systems.
Learned behaviors can also be brittle, not generalizing to minor changes in the environment or plant being controlled, which can lead to overall poor performance. To address these challenges, this work proposes a human-in-the-loop methodology for learning and adapting a control policy in an online fashion. In particular, the approach learns the control policy from a human-demonstrated task and adapts it based on human interventions, while the learned controller is performing the task. 
Because the learning and adaptation of the model occur online with this method, learning is faster, physically validated, and generally needs less training data.

The idea of learning from human demonstrations and human interventions has previously been explored in \cite{Ross2013} to re-label undesired behavior, to preemptively intercede in avoiding catastrophic failure \cite{Hilleli2018} and learn from these interventions \cite{Goecks2019}, to guide learning through sparse evaluative feedback indirectly \cite{Warnell2018}, and to combine these human modalities with reinforcement learning \cite{Waytowich2018,goecks2019integrating}, even when the human demonstrations and interventions were collected from both proficient and non-proficient users who had to decide when to intervene \cite{WU2022}.
However, a human-in-the-loop machine learning framework has not been applied to cases where the task complexity changes and real-time human interventions are model-uncertainty informed.




In this paper, real-time human demonstrations and interventions informed by uncertainty are used to train a deep neural network-based quadrotor flight controller to land on static and dynamic landing platforms.
The quadrotor is first trained to land on a static platform with only human demonstrations and compared to classic controllers performing the same task.
It is then shown that the model performance can improve if retrained with additional data from real-time interventions that correct the system behavior when the model's uncertainty is above a predefined threshold.
The neural network uncertainty is computed via Monte Carlo Dropout \cite{pmlr-v48-gal16} and the human is signaled to intervene by an audiovisual cue when supervising the vehicle's behavior.
Finally, the task changes to landing on a moving platform where experiments show that a small number of interventions are sufficient to adapt the model in real-time to changes in the task, as opposed to classic controllers that need to be redesigned offline.
We show that, by using this human-in-the-loop machine learning framework, human-like control policies can be augmented with minimal training time and can succeed without further hand-tuning.
All presented machine learning routines are executed onboard the vehicle.

The main contributions of this work are:

\begin{itemize}
    \item Novel human-in-the-loop machine learning framework that uses the estimated model uncertainty for targeted training to improve task execution;
    \item Experiments comparing uncertainty-informed human-in-the-loop machine learning technique to classic controllers, showing its advantage when working with dynamic tasks;
    \item Extensive hardware validation of a novel human-in-the-loop machine learning technique that learns tasks in real-time \cite{Goecks2019}.
\end{itemize}

\section{Background }





In this work, we represent our learning controllers with deep neural networks.
Although deep neural network techniques have shown great success, they often require large amounts of data and lengthy training times to converge to good solutions.
Human-in-the-loop machine learning is a class of techniques that augment this training process with data collected from various forms of human interaction. 

\subsection{Learning from Human Demonstrations}

\begin{figure}[thpb]
  \centering
  \includegraphics[width=0.85\columnwidth]{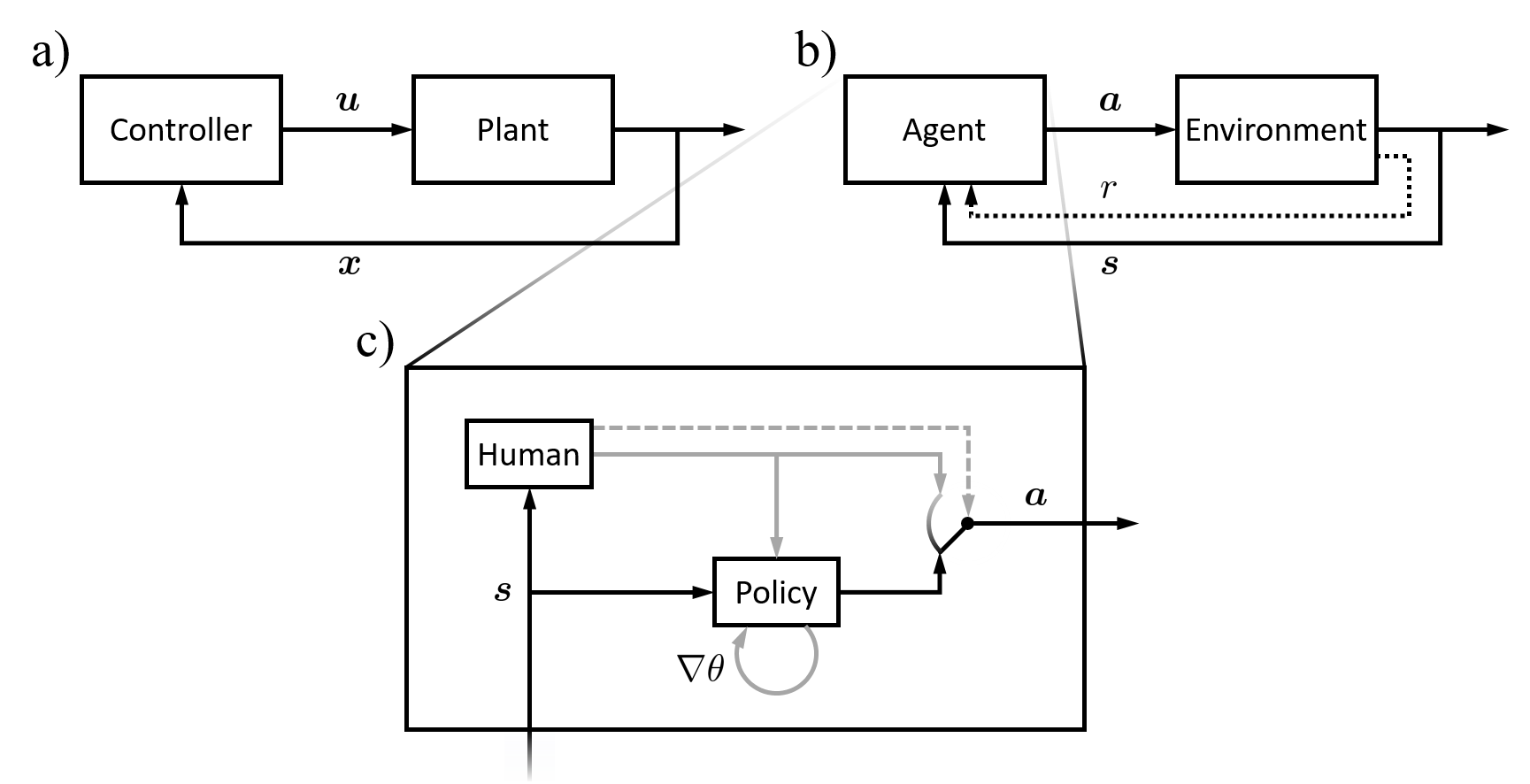}
  \caption{Block diagrams for a) typical feedback control systems and b) analogous learning agent systems with reward signal (dotted). c) Inset diagram of the \emph{Agent} block for LfD and LfI, where the human triggers (dashed) switch between actions output by the current policy (black) and actions output from themselves with simultaneous policy update (gray).}
  \label{f:blk_diag}
\end{figure}

\emph{Learning from Demonstration} (LfD) leverages human demonstrations or examples of the task for learning \cite{Argall2009,Osa2018AnAP}. 
In LfD, the policy $\pi_\theta$ is a function parameterized by $\theta$ that maps from states $\vecb{s}$ to actions $\vecb{a}$.
\begin{equation} \label{eq:policy}
\pi_\theta : S \rightarrow A
\end{equation}
In control theory notation, $\vecb{s}$ would be analogous to the system state vector $\vecb{x}$, and $\vecb{a}$ would correspond to the control vector $\vecb{u}$, as shown in Fig. \ref{f:blk_diag}a-b.
For the case of partial observability, observation $\vecb{o}$ would be used in place of state $\vecb{s}$, corresponding to the measurement vector $\vecb{y}$ used in control theory for systems without full-state feedback.

The policy $\pi_\theta$ is trained to generalize over a set $\mathcal{D}$ of states and actions visited during a task demonstration spanning $T$ time steps, i.e. a time history of the state and action trajectories:
\begin{equation} \label{eq:demonstration_dataset}
\mathcal{D} = \left\{\vecb{a}_1, \vecb{s}_1, \vecb{a}_2, \vecb{s}_2, ... , \vecb{a}_T, \vecb{s}_T\right\}.
\end{equation}
This demonstration may be performed by an analytically derived controller, a pre-trained policy, or a human supervisor.
In the case of human demonstrations, the human provides the actions in response to observed states, subject to what may be represented as their internal policy $\pi_{sup}$, thereby generating $\mathcal{D}$. 
The training of policy $\pi_\theta$ takes the form of a standard supervised learning problem to estimate the human's internal policy as $\hat{\pi}_{sup}$, where the parameters $\theta$ are trained in order to minimize a loss function, such as mean squared error, as shown in (\ref{eq:sup_mse}).


\begin{equation}\label{eq:sup_mse}
\hat{\pi}_{sup} = \argmin_{\pi_\theta} \frac{1}{T}\sum_{t=1}^{T} ||\pi_\theta(\vecb{s}_t) - \vecb{a}_t ||^2
\end{equation}

Several recent empirical successes have been demonstrated in applying imitation learning to train autonomous systems in the domains of self-driving cars \cite{Bojarski2016} and simulated \cite{Goecks2018} and real drone flight \cite{Giusti2015}.



\subsection{Learning from Human Interventions}

In \emph{Learning from Intervention} (LfI) the human acts as a supervisor, observing the agent performing the task and intervening, that is overriding agent actions, when necessary to avoid undesirable or unsafe behaviors that may lead to catastrophic states. 
This technique was used to train agents without catastrophic failures, in a safe reinforcement learning framework \cite{Saunders2017}, where a human overseer monitored a learning agent and provided inputs that blocked the actions of the agent when the human deemed them to be potentially catastrophic. This data was then used to train a function to perform this blocking automatically.
Similar work by \cite{Hilleli2018} used human interaction to train a classifier for detecting unsafe states.


\subsection{Learning from Demonstrations and Interventions}

Combining LfD and LfI has been successfully demonstrated \cite{Goecks2019} to train a fully connected deep neural network to land a drone on a target landing pad in a high-fidelity drone simulator, and an autonomous driving task \cite{hgdagger2019} in both simulated and real test vehicles.
Both previous works demonstrated that the combination of these learning techniques outperformed either technique in isolation and with greater data efficiency and is the main inspiration for this work.

The users giving the interventions, however, had to rely on their own instincts to decide when to correct the agent's behavior, which adds to the cognitive burden already imposed on the pilot. 
In this work, we alleviate that burden by computing in real-time and displaying to the human the model's uncertainty as an indicator of when to intervene.

\section{Methodology}

The proposed method starts by collecting a set of task demonstrations from an expert human demonstrator, which, in this case, entails landing the quadrotor on a fixed platform.
A policy $\pi_\theta$ is initially trained using this set of task demonstrations according to Equation \ref{eq:sup_mse}.
This policy takes as input the ground truth relative $x$ and $y$ positions between the quadrotor and the landing pad, the ground truth global yaw of the vehicle in Euler angle, and the estimated $z$ position (altitude) of the vehicle computed using the open-source REEF estimator \cite{Ramos2019REEF}.
The outputs of the policy are velocities generated from joystick inputs controlled by the human demonstrator.
These reference velocities are in turn sent to the REEF Adaptive Control that outputs \textit{torque} commands to meet the desired velocities.
The role of this adaptive controller is to handle minor changes in vehicle dynamics, such as changes in the trim or battery configurations, and improve flight performance.

After the policy is trained with this initial set of task demonstrations, the agent is given control and executes $\pi_\theta$ while the human supervises the agent's actions. 
While the agent performs the task, we compute the neural network's uncertainty using Monte Carlo Dropout \cite{pmlr-v48-gal16}, a technique where dropout layers are used during inference to produce a number of stochastic forward passes, which are then aggregated to compute the predictive mean and predictive uncertainty of the neural network model. 
The main advantage of this method is that it can be applied to already existing neural network models trained with dropout layers, as long as these layers can be enabled during inference.

The estimated uncertainty values are converted to an audiovisual signal that cues the human supervisor to intervene when the agent is navigating regions of high uncertainty.
The human also intervenes whenever the agent exhibits unwanted behavior, provides corrective actions, and then releases control back to the agent when they deem it appropriate. 
The agent then learns from this intervention by augmenting the original training dataset with the states and actions recorded during the intervention, fine-tuning $\pi_\theta$. 
The agent then executes the new policy $\pi_\theta$, as it is updated after the interventions are collected, with the human user continuing to supervise and intervene as needed, toggling control of the vehicle between themselves and the current agent policy via a button on the joystick.

The human user can fluidly switch at will between providing demonstrations, i.e. uninterrupted state-action trajectories for an entire landing maneuver, and interventions, i.e. corrective state-action trajectories applied to segments of the landing maneuver,  as seen in Fig. \ref{f:blk_diag}c.
By combining demonstrations and interventions, this framework can outperform a policy learned from human demonstrations alone, and with less training data \cite{Goecks2019}. 
This effect is due in part to the more diverse exploration of the state space seen during the LfI stage than was experienced in the LfD stage, which would have resulted in failure. 
Targeted intervention during these failure states allows the agent to efficiently update the policy in its most deficient regions.
Once the agent is proficient in landing on the fixed platform, the task slowly changes by adding motion to the landing pad.
Once the platform is in motion, more human interventions are collected in order to update the policy $\pi_\theta$ in real-time to this new task.

\subsection{Implementation Details}

The learning framework used in this work is summarized in Algorithm \ref{a:lfd_lfi}. 
The main procedure initializes the agent's policy $\pi_\theta$, the human dataset $\mathcal{D}$, and a secondary \emph{Policy Update} routine to update the policy with the new demonstration or intervention data. 
The agent's policy $\pi_\theta$ is a fully-connected, three hidden-layer, neural network with 130, 72, and 40 neurons, respectively. 
All layers are fully-connected with hyperbolic tangent functions as activation.
Dropout layers \cite{srivastava2014dropout} with 25\% drop probabilities are used in the first two layers to prevent overfitting and compute uncertainty.
The network is randomly initialized with weights sampled from a normal distribution. 

The main loop primarily comprises of reading the current environment state and executing actions provided by either the agent's current policy or the human supervisor.
When the human supervisor interacts by holding a trigger on the joystick and providing input flight commands, these actions are executed by the agent and stored with the current state in the dataset $\mathcal{D}$.
This joystick trigger is used to toggle control authority between the human and the current policy.
If the human is not depressing this joystick trigger, multiple actions $\vecb{a}$ are sampled from the current policy $\pi_\theta$, given the current state $\vecb{s}$, and the mean value is executed by the agent.
Action standard deviation is used as a proxy for the model uncertainty to guide the data collection via human interventions.

In the secondary \emph{Policy Update} procedure, the maximum number of training epochs $n_{max}$ and the batch size $N$ are initialized.
The procedure performs policy updates when the human user provides either demonstration or intervention data. 
It loads the current dataset $\mathcal{D}$ and checks for newly added data, based on the number of samples.
If new samples are found, this loop samples a batch of $N=64$ state-action pairs from $\mathcal{D}$. 
Given these states $\vecb{s}$, corresponding actions $\hat{\vecb{a}}$ are sampled from the current policy $\pi_\theta$, and the mean squared error loss $l$ is computed:
\begin{equation}
    l = \frac{1}{N} \sum_{i=1}^{N} (\hat{\vecb{a}}_i - \vecb{a}_i)^2.
\end{equation}
If the number of optimization epochs is below the maximum $n_{max}=30$, the policy $\pi_\theta$ is optimized with the commonly used Root Mean Square Propagation (RMSProp) optimizer, with a learning rate of 1e-4. 

\begin{algorithm}[!tb]
\caption{Learning from Human Demonstrations and Uncertainty-informed Interventions}\label{a:lfd_lfi}
\begin{algorithmic}[1]

\Procedure{Main}{}
    \Initialize{policy $\pi_\theta$, dataset $\mathcal{D}$}
    \While {True}
        \State Record state $\vecb{s}$
        \If {Human interacting}
            \State Record human action $\vecb{a}$
            \State Add state-action pair $\vecb{s}$, $\vecb{a}$ to $\mathcal{D}$
        \Else
            \State Sample multiple actions $\vecb{a}$ from policy $\pi_\theta(\vecb{s})$
            \State Compute mean and standard deviation of the actions
            \State Use mean as the selected action $\vecb{a}$ 
            \State Use standard deviation as the model uncertainty
        \EndIf
        \State Execute action $\vecb{a}$
        \If {Task completed} 
            \State Reset task and read initial state $\vecb{s}_0$
            \Initialize{procedure \textsc{Policy Update}}
            \State \Return Initial state $\vecb{s}_0$
        \EndIf
    \EndWhile
\EndProcedure

\Procedure{Policy Update}{}
    \Initialize{epochs $n_{max}$, batch size $N$}
    \While {True}
        \State Load human dataset $\mathcal{D}$
        \If {New samples in $\mathcal{D}$}
            \While {$n < n_{max}$}
                \State Sample $N$ state-action pairs ($\vecb{s}, \vecb{a}$) from the expert dataset $\mathcal{D}$
                \State Sample action $\hat{\vecb{a}}$ from policy $\pi_\theta(\vecb{s})$
                \State Compute loss $l$
                \State Increment epoch $n$
                \State Perform gradient descent update on $\pi_\theta$
            \EndWhile
        \EndIf
    \EndWhile
\EndProcedure

\end{algorithmic}
\end{algorithm}

\section{Hardware Results}
This section describes the experimental setup used to execute the proposed framework on hardware platforms and presents the results from the experiments. 

\subsection{Experimental Setup}

\begin{figure}[thpb]
  \centering
  \includegraphics[width=1.0\columnwidth]{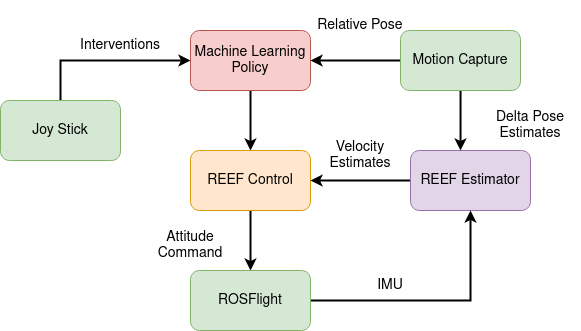}
  \caption{Block diagram depicting the overall system used for demonstration, intervention, and deployment of the learned policy $\pi_{\theta}$ within the control framework of the quadrotor.}
  \label{f:block_diagram}
\end{figure}

A human operator provides the demonstrations and interventions for the machine learning model by commanding body-level (forward and lateral) velocities, yaw rate, and altitude commands from a Logitech Gamepad controller. These commands are then fed into a velocity and altitude controller, which then sends attitude and thrust commands to a flight controller running ROSFlight \cite{jackson2016rosflight}. In this landing task, we use the REEF PID and the REEF Adaptive controllers \cite{Ramos2019REEF} to compensate for minor changes in the multi-rotor trim setting. The REEF controllers are open-source flight controllers developed and maintained by the Research and Education Engineering Facility (REEF) at the University of Florida.
Fig. \ref{f:block_diagram} shows the block diagram of the overall experimental system. A detailed implementation of the flight controllers can be found at the GitHub page\footnote{REEF Adaptive Controller: \url{https://github.com/uf-reef-avl/reef_adaptive_control}}.

The inputs to the neural networks are the relative positions of the quadrotor with respect to the target in the camera frame. This measurement is obtained from a motion capture system that tracks the quadrotor and the landing pad positions. An extrinsic calibration is performed using \cite{ganesh2021extrinsic}, which enables motion capture measurements as a stand-in for onboard measurements, such as using ArUco markers.
In addition to the relative positions, the neural network receives the heading and altitude of the quadrotor, obtained from motion capture and REEF Estimator, respectively. Once the landing demonstrations are recorded, the network is trained and deployed (also referred to as a \emph{rollout}) using the same framework shown in Fig.~\ref{f:block_diagram}.

\begin{figure}[thpb]
  \centering
  \includegraphics[width=.85\columnwidth]{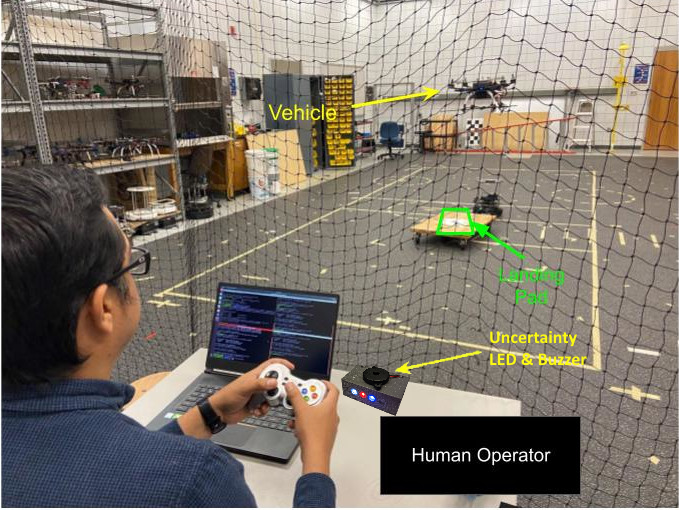}
  \caption{The human operator intervening to correct undesirable actions of the quadrotor to improve the landing task performance. The landing pad being towed by a ground robot is also seen in the image.}
  \label{f:col_human}
\end{figure}

For each rollout, the human supervisor is guided to make the decision to intervene based on an audiovisual representation of the current model's uncertainty (Fig.~\ref{f:col_human}). Effectively, as the estimated uncertainty value exceeds user-defined thresholds, an accumulative number of lights are turned on commensurate to the model uncertainty. Based on these signals, the human supervisor is informed about how critical the intervention can be. In addition, when the uncertainty is beyond a maximum threshold, a buzzer will beep to suggest that the model is unreliable and the human must intervene. All uncertainty thresholds are set heuristically based on the estimated uncertainties obtained from the demonstration flights. This uncertainty is estimated by the variability of the neural network which is quantified by the standard deviation induced by Monte Carlo Dropout. As the human supervisor intervenes, each of these interventions is recorded and used to fine-tune the agent's behavior. 

\begin{figure}[thpb]
  \centering
  \includegraphics[width=1.0\columnwidth]{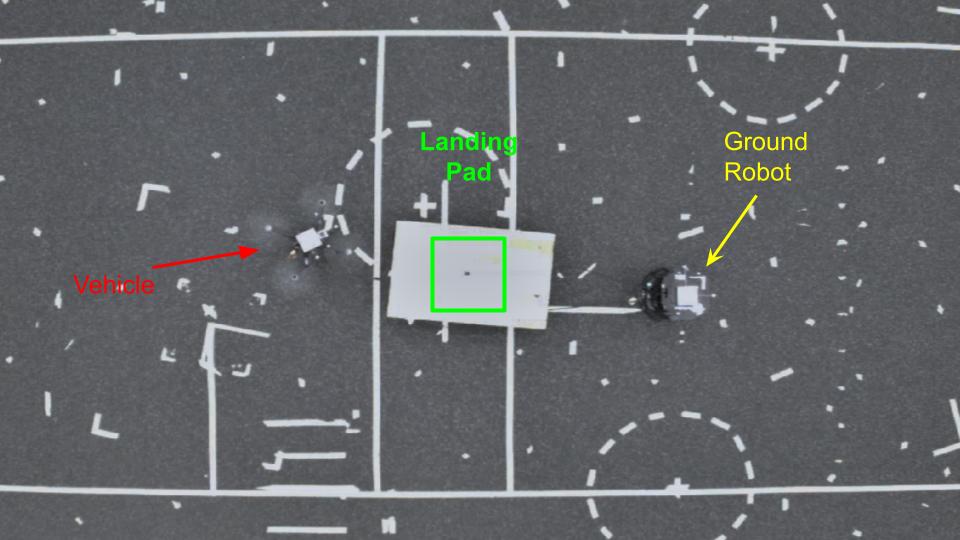}
  \caption{A typical experiment captured by an overhead camera at the lab space. The image shows the quadrotor vehicle following the landing pad which is being towed by the ground robot. An ArUco marker is used to denote the center of the landing pad.}
  \label{f:experimental_setup}
\end{figure}

In order to demonstrate that the proposed framework can easily handle changes in the environment or dynamics, the landing task is performed on a fixed, and then a moving pad.
The experimental setup, as seen in Fig.~\ref{f:experimental_setup}, comprises a ground robot towing the landing pad on a cart.
The linear velocity of the landing pad is approximately 0.25m/s. 

\subsection{Hardware Results}
The hardware results are presented in the following subsections: \emph{Demonstration and Rollout}, where human demonstrations are used to train the network to imitate the landing task on a static landing pad; \emph{Uncertainty-informed Interventions}, where interventions are used to reduce the network uncertainty and fine-tune its performance; and \emph{Landing on a Moving Platform}, where interventions are used to adapt the learned model to changes in the environment. In this case, to adapt the learned landing to handle a moving landing pad.  

\subsubsection{Demonstrations and Rollout}

\begin{figure}[htbp]
    \centering
    \includegraphics[width=0.85\columnwidth]{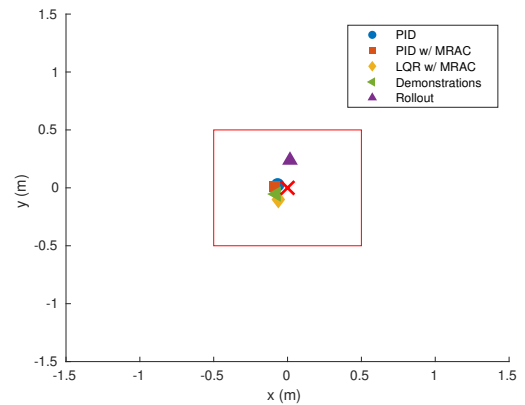}
    \caption{The mean landing location of the quadrotor under different control policies is shown relative to a square landing area representing the size of the quadrotor base (approximately 50 cm by 50 cm). The center of the landing pad is denoted with a cross. The demonstrations have a comparable mean landing position as that of the classical controllers with a mean radial error of 12.73 cm. The mean landing location of the initial rollout is skewed from the demonstrations due to model uncertainty.  }
    \label{fig:mean_landing}
\end{figure}

In order to execute the landing task, the network utilizes a training set of 10 human demonstrations. For each demonstration, the quadrotor begins the learning experiment from a different initial position to diversify the training data. During the demonstrations, human inputs, i.e., velocities, altitude, and yaw rate, are saved along with the network inputs.
Then, the network is trained for 30 epochs and deployed from a uniform selection of initial positions, distinct from the takeoff position from the learning set, to execute the landing task.  Fig.~\ref{fig:mean_landing} shows the mean landing location  of the quadrotor when starting from 6 unique takeoff locations. The mean landing position for human demonstrations and classical controllers is also shown in this figure. From the figure, it is also observed that the network with only demonstrations is able to imitate the human pilot and successfully land the quadrotor within the landing pad, near the center but deviated from the demonstration mean landing point.

\begin{figure}[thpb]
  \centering
  \includegraphics[width=.85\columnwidth]{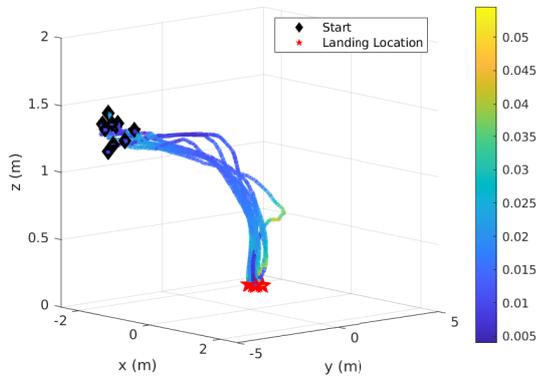}
  \caption{Figure showing the trajectory of the quadrotor executing the landing task from the same starting location. The trajectories are colored based on the uncertainty from the model trained with 10 demonstrations similar to that of the stationary platform landing imitation using multiple takeoff locations presented earlier.}
  \label{f:uncertainty_plot}
\end{figure}

The 10 landing trajectories produced by the network when trained with demonstrations only are shown in Fig.~\ref{f:uncertainty_plot}. The landing trajectories are shaded based on the model uncertainty obtained. A higher number indicates higher model uncertainty. In particular, the color bar indicates the sum of variance in each control axis (forward and lateral velocities, and altitude). While the landing task follows similar trajectories when starting approximately from the same location, the uncertainty metric does capture the output variability.This was seen especially during the descending stage, where higher variations in relative position results are present due to ground effects and final corrective commands from the demonstrations.

\subsubsection{Uncertainty-informed Interventions}

Since the learning is performed in real-time, the decision to intervene can also be made in real-time, periodically, and on demand. To decide if an intervention is needed, the human operator gives control to the agent and observes if the flight path generated is appropriate and if the network uncertainty may lead to an unsafe or erratic flight.
To estimate the network uncertainty, we use the standard deviation of a set of network outputs induced by Monte Carlo Dropout. In particular, the current input is fed forward through the network 1000 times, each time dropping a random set of network neurons. By doing so, 1000 outputs per input are produced; the sampled standard deviation of these 1000 outputs is what we consider the estimated uncertainty of the model. This computation is performed online to inform the pilot when they should intervene. As previously mentioned, the pilot is notified when the model uncertainty is high, exceeding aforementioned thresholds, with the help of a buzzer and LED lights, as seen in Fig. \ref{f:col_human}. 
Note that in the event the controller performs an unexpected flight maneuver, the human can also intervene and override control inputs via manual remote commands directly to the quadrotor torques. 

\begin{figure}[thpb]
  \centering
  \includegraphics[width=.85\columnwidth]{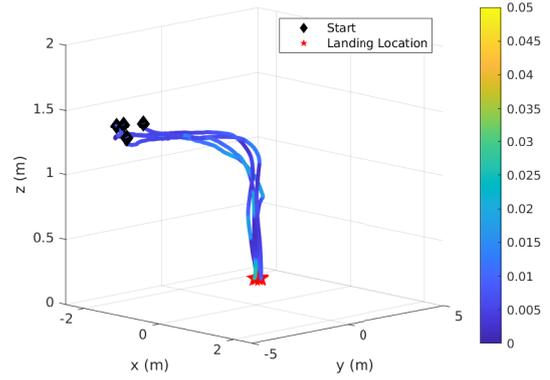}
  \caption{Trajectory of the quadrotor after fine-tuning the policy with uncertainty-informed human interventions. The uncertainty and the final landing locations improve based on this intervention.}
  \label{f:post_intervention}
\end{figure}

\begin{figure}[thpb]
    \centering
    \includegraphics[width=0.85\columnwidth]{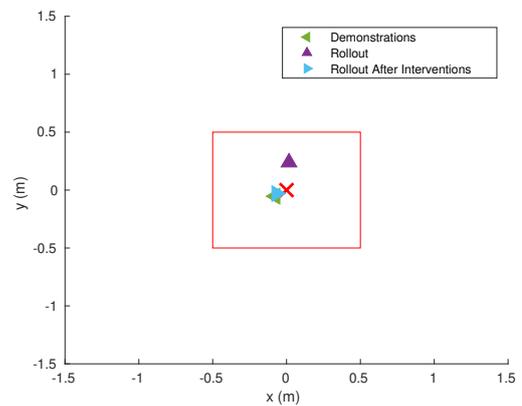}
    \caption{The mean landing location of the quadrotor under the learned control policies, highlighting the effect of the interventions. The interventions improve the rollout of the policy so that the mean landing location mimics that of the initial demonstrations.}
    \label{fig:mean_landing_intervene}
\end{figure}

The interventions were carried out when the model was rolled out from 4 unique starting locations. For each rollout, the pilot decided when to intervene based on the audiovisual cues representing the model's uncertainty.
Fig.~\ref{f:post_intervention} shows the landing trajectories for the landing task after incorporating the uncertainty-informed interventions into the model.
Fig.~\ref{fig:mean_landing_intervene} shows the mean landing locations for these rollouts.

From Fig.~\ref{f:post_intervention} and Fig.~\ref{fig:mean_landing_intervene}, one can note that the quadrotor trajectories present less variability after the interventions, and the overall accuracy of the quadrotor mimics the demonstrated trajectories better. Quantitatively the improvement of the landing task after interventions is presented in Table ~\ref{table:mean_variance} using mean landing location errors and deviation in the cartesian and radial directions. The intervention-trained policy has a landing accuracy comparable to the classical controllers and demonstrations. However, it performs with superior precision. Note that the model uncertainty was also positively impacted, being reduced by a factor of 5 with respect to the performance of the demonstration-only training. Also, recall that the only additional information is from interventions at times where the audiovisual device signaled above-threshold uncertainties.

\begin{table}[thpb]
\caption{Mean and deviation of landing locations for each control policy.}
\label{table:mean_variance}
\centering

\begin{tabular}{|c| c c c|}
    \hline
        \text{PID} & x (cm) & y (cm) & r (cm) \\
        \hline
        $\mu$ & -6.73 & 2.57 & 17.99 \\
        $\sigma$ & 15.35 & 11.39 & 6.27 \\ 
        \hline
        
        \text{MRAC} & x (cm) & y (cm) & r (cm) \\
        \hline
        $\mu$ & -9.03 & 0.73 & 11.74 \\
        $\sigma$ & 10.85 & 5.34 & 8.92 \\ 
        \hline

        \text{LQR} & x (cm) & y (cm) & r (cm) \\
        \hline
        $\mu$ & -6.23 & -10.08 & 15.08 \\
        $\sigma$ & 5.91 & 12.00 & 8.64 \\ 
        \hline

        \text{Demo.} & x (cm) & y (cm) & r (cm) \\
        \hline
        $\mu$ & -7.61 & -5.34 & 12.73 \\
        $\sigma$ & 5.53 & 8.48 & 4.30 \\ 
        \hline

        \text{Rollout} & x (cm) & y (cm) & r (cm) \\
        \hline
        $\mu$ & 1.85 & 23.83 & 26.82 \\
        $\sigma$ & 3.51 & 18.56 & 13.35 \\ 
        \hline

        \text{R.A.I.} & x (cm) & y (cm) & r (cm) \\
        \hline
        $\mu$ & -6.82 & -3.02 & 5.26 \\
        $\sigma$ & 6.28 & 8.83 & 1.10 \\ 
        \hline
\end{tabular}
\end{table}

\subsubsection{Landing on a Moving Platform}
The proposed framework allows the incorporation of human interventions to adapt the model behavior to related, although significantly different, tasks. To demonstrate this concept, the model used for landing on a stationary target is adapted to landing on a moving target by learning from human interventions. 

\begin{figure}[H]
    \centering
    \includegraphics[width=0.95\columnwidth]{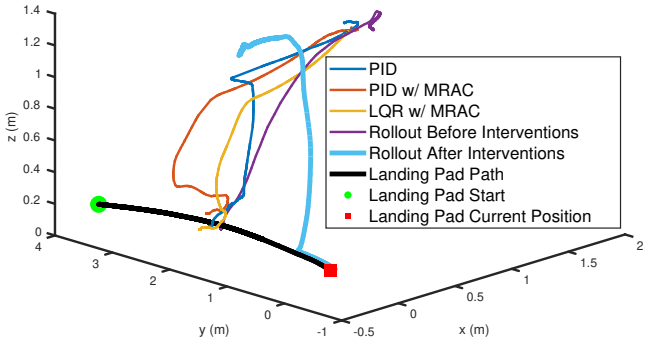}
    \caption{Attempts to land on a moving target under multiple control policies. Only the network policy after interventions is capable of landing on the target. The green circle indicates the direction of travel of the landing pad at the start, and a red square at the end of the trajectory.}
    \label{fig:moving_landing}
\end{figure}

Fig.~\ref{fig:moving_landing} shows the performance of the policy before and after interventions, along with that of the classical controllers, when landing on a moving target. All classical controllers, along with the policy $\pi_\theta$ trained only on the stationary target, fail to land on the dynamic platform.

To adapt the model behavior, the policy $\pi_\theta$ is retrained after three human interventions to incorporate the information of a moving landing pad. The network was trained with these three additional interventions for 10 epochs. The improved policy, shown in blue in Fig.~\ref{fig:moving_landing}, successfully learned the new dynamics of the dynamic landing platform and was able to land on the moving target.


Note that landing on a moving platform is just an example of the task/environment being changed. The same framework can be used to execute other tasks which are relatively more complicated, like flying around an obstacle, executing certain landing maneuvers (like a cork-screw landing), or even flying through windows without needing rigorous models.



\section{Conclusions}

In this work, we presented a novel human-in-the-loop machine learning framework that uses uncertainty estimation for targeted training of deep neural network-based controllers to improve task execution.
Our proposed methodology successfully combined learning from human demonstrations with uncertainty-informed interventions to teach a small unmanned aerial vehicle to land on a fixed landing pad and, by augmenting the learned task using uncertainty-informed interventions, to land on a moving landing platform. 
We showed hardware validation experiments comparing our uncertainty-informed human-in-the-loop machine learning technique to classic controllers. The results demonstrate that uncertainty-informed interventions, in addition to human demonstration-based learning, are advantageous to classical methods when working with dynamic tasks.
This adaptability of a deep neural network via human interventions has the potential to enable robots to adapt to changing environments and learn new tasks online with goals aligned with human objectives. 


\bibliography{root.bib}             
\end{document}